\PassOptionsToPackage{table,xcdraw}{xcolor}  
\documentclass[sigconf]{acmart}
\AtBeginDocument{%
	\providecommand\BibTeX{{%
			\normalfont B\kern-0.5em{\scshape i\kern-0.25em b}\kern-0.8em\TeX}}}

\setcopyright{acmcopyright}

\copyrightyear{2025}
\acmYear{2025}
\setcopyright{acmlicensed}\acmConference[MM '25]{Proceedings of the 33rd
ACM International Conference on Multimedia}{October 27--31, 2025}{Dublin,
Ireland}
\acmBooktitle{Proceedings of the 33rd ACM International Conference on
Multimedia (MM '25), October 27--31, 2025, Dublin, Ireland}
\acmDOI{10.1145/3746027.3755868}
\acmISBN{979-8-4007-2035-2/2025/10}

\usepackage{booktabs}           
\usepackage{multirow}           
\usepackage{array}              
\usepackage{xcolor}
\usepackage{tabularx}   
\usepackage{subcaption}  
\usepackage{pifont}
\usepackage{makecell}

\definecolor{green}{HTML}{D1E9D3}

\definecolor{MyGreen}{HTML}{61A64C}
\definecolor{MyYellow}{HTML}{FFD13F}

\newcommand{\cmark}{\textcolor{MyGreen}{\ding{51}}} 
\newcommand{\xmark}{\textcolor{MyYellow}{\ding{55}}} 

\begin{document}

\title{SafeDriveRAG: Towards Safe Autonomous Driving with Knowledge Graph-based Retrieval-Augmented Generation}

\author{Hao Ye}
\affiliation{%
  \institution{State Key Laboratory of Networking and Switching Technology, Beijing University of Posts and Telecommunications}
  \city{Beijing}
  \country{China}
}
\email{haoye@bupt.edu.cn}
\orcid{0009-0002-4738-4139}
\author{Mengshi Qi}
\authornote{Corresponding author}
\affiliation{%
  \institution{State Key Laboratory of Networking and Switching Technology, Beijing University of Posts and Telecommunications}
  \city{Beijing}
  \country{China}
}
\email{qms@bupt.edu.cn}
\orcid{0000-0002-6955-6635}
\author{Zhaohong Liu}
\affiliation{%
  \institution{State Key Laboratory of Networking and Switching Technology, Beijing University of Posts and Telecommunications}
  \city{Beijing}
  \country{China}
}
\email{liuzhaoh@bupt.edu.cn}
\orcid{0009-0000-2066-0215}
\author{Liang Liu}
\affiliation{%
  \institution{State Key Laboratory of Networking and Switching Technology, Beijing University of Posts and Telecommunications}
  \city{Beijing}
  \country{China}
}
\email{liangliu@bupt.edu.cn}
\orcid{0009-0003-2173-9226}
\author{Huadong Ma}
\affiliation{%
  \institution{State Key Laboratory of Networking and Switching Technology, Beijing University of Posts and Telecommunications}
  \city{Beijing}
  \country{China}
}
\email{mhd@bupt.edu.cn}
\orcid{0000-0002-7199-5047}
\renewcommand{\shortauthors}{Hao Ye, Mengshi Qi, Zhaohong Liu, Liang Liu, \& Huadong Ma}

\begin{abstract}
In this work, we study how vision-language models (VLMs) can be utilized to enhance the safety for the autonomous driving system, including perception, situational understanding, and path planning. However, existing research has largely overlooked the evaluation of these models in traffic safety-critical driving scenarios. To bridge this gap, we create the benchmark (SafeDrive228K) and propose a new baseline based on VLM with knowledge graph-based retrieval-augmented generation (SafeDriveRAG) for visual question answering (VQA). Specifically, we introduce SafeDrive228K, the first large-scale multimodal question-answering benchmark comprising 228K examples across 18 sub-tasks. This benchmark encompasses a diverse range of traffic safety queries, from traffic accidents and corner cases to common safety knowledge, enabling a thorough assessment of the comprehension and reasoning abilities of the models. Furthermore, we propose a plug-and-play multimodal knowledge graph-based retrieval-augmented generation approach that employs a novel multi-scale subgraph retrieval algorithm for efficient information retrieval. By incorporating traffic safety guidelines collected from the Internet, this framework further enhances the model’s capacity to handle safety-critical situations. Finally, we conduct comprehensive evaluations on five mainstream VLMs to assess their reliability in safety-sensitive driving tasks. Experimental results demonstrate that integrating RAG significantly improves performance, achieving a +4.73\% gain in Traffic Accidents tasks, +8.79\% in Corner Cases tasks and +14.57\% in Traffic Safety Commonsense across five mainstream VLMs, underscoring the potential of our proposed benchmark and methodology for advancing research in traffic safety. Our source code and data are available at https://github.com/Lumos0507/SafeDriveRAG.
\end{abstract}



\maketitle

\section{Introduction}

In recent years, Vision-Language Models~\cite{liu2023visual,qwen2vl2024,yu2024rlaif}~(VLMs) have attracted considerable attention in the field of autonomous driving. Numerous studies~\cite{sima2025drivelm,mao2024language,malla2023drama} have explored to integrate VLMs into autonomous driving systems to enhance perception and decision-making capabilities in conventional traffic environments, demonstrating impressive performance across a range of general tasks. However, these existing efforts primarily focus on standard driving scenarios, while systematically overlooking the safety performance of VLMs under non-standard conditions, such as traffic accidents~\cite{zheng2023avoid} and corner cases~\cite{breitenstein2021corner}. Given that safety is one of the most critical metrics~\cite{WHO2023RoadSafety} for end-to-end autonomous driving, an inadequate focus on safety evaluation could result in system failures at key moments, leading to severe consequences. Consequently, whether VLMs can adequately understand and respond to complex traffic incidents and simultaneously make effective decisions to ensure the safety of driving decisions remains an urgent yet challenging issue.

\begin{table*}[t]
\small
\centering
\caption{Comparison of existing multimodal driving datasets in terms of QA types (Single or Multiple choice), question categories (accident, corner case, safety commonsense, laws and regulation) and data size.} 
\vspace{-3mm}
\begin{tabular}{l|cc|cc|cccc|c}
\toprule
 \multirow{3.5}{*}{\textbf{Dataset}} & \multirow{3.5}{*}{\textbf{Venue}} &  \multirow{3.5}{*}{\textbf{Tasks} }
& \multicolumn{2}{c|}{\textbf{QA Types}} 
& \multicolumn{4}{c|}{\textbf{Question Categories}} 
&  \multirow{3.5}{*}{\textbf{Size}} \\
\cmidrule(lr){4-5} \cmidrule(lr){6-9}
& & 
& \textbf{Single} & \textbf{Multiple} 
& \textbf{Accident} & \textbf{Corner Case} 
& \textbf{\makecell{Safety\\Commonsense}} 
& \textbf{Laws \& Reg.} 
& \\
\midrule
BDD-X\cite{kim2018textual}         & ECCV18   & Video QA         & \xmark & \xmark & \xmark & \xmark & \cmark & \xmark & 26K \\
DRAMA\cite{malla2023drama}         & WACV23   & Video QA         & \xmark & \xmark & \cmark & \xmark & \cmark & \xmark & 102K \\
DriveLM\cite{sima2025drivelm}       & ECCV24   & Image QA         & \cmark & \xmark & \xmark & \xmark & \cmark & \xmark & 2M \\
nuScenes-QA\cite{qian2024nuscenes}   & AAAI24   & Image QA         & \cmark & \xmark & \xmark & \xmark & \xmark & \xmark & 460K \\
CODA-LM\cite{chen2024automated}       & WACV25   & Image QA         & \cmark & \xmark & \cmark & \xmark & \xmark & \xmark & 60K \\
IDKB\cite{lu2024lvlms}          & AAAI24   & Image QA         & \cmark & \cmark & \xmark & \xmark & \cmark & \cmark & 1M \\
\textbf{SafeDrive228K} & --       & Video QA, Image QA & \cmark & \cmark & \cmark & \cmark & \cmark & \cmark & 228K \\
\bottomrule
\end{tabular}
\label{tab:dataset_comparison}
\end{table*}

To address this issue, numerous efforts have been made to build visual-language datasets ~\cite{sima2025drivelm,malla2023drama,lv2025t2sg,lu2024lvlms,chen2024automated,karim2023attention,qian2024nuscenes,tian2024drivevlm} tailored for complex driving scenarios. However, these existing datasets generally concentrate on a specific subdomain such as perception and recognition~\cite{qian2024nuscenes}, extreme weather conditions~\cite{Marathe_2023_CVPR}, corner-case scenarios~\cite{li2022coda,wen2023roadgpt4v}, or particular driving skills~\cite{lu2024lvlms}. Despite the recent growth of multimodal datasets related to driving safety, a unified benchmark capable of systematically and comprehensively evaluating models’ performance in safe driving remains lacking. 

To this end, we propose a new benchmark SafeDrive228K designed to comprehensively evaluate VLMs ability to understand and respond in diverse safety-critical driving situations. This benchmark consists of three sets of sub-tasks: (1) \emph{Traffic Accident Tasks}, containing 10K real-world traffic accident videos along with 102K structured question-answer pairs for assessing a model’s capabilities in recognizing, analyzing, and managing actual accident scenarios; (2) \emph{Corner Case Tasks}, comprising 10K images of real-world corner cases and 69K structured question-answer pairs to evaluate the model’s robustness in handling rare or intricate traffic contexts; and (3) \emph{Traffic Safety Commonsense Tasks}, featuring 26K images and 57K structured question-answer pairs that span common safety issues in everyday driving, thereby testing the model’s fundamental knowledge of driving safety.

Then, we conduct the systematic evaluation of several representative open-source VLMs, including Qwen2.5-vl~\cite{bai2025qwen25vl}, LLAVAA~\cite{liu2023visual}, and Phi-4~\cite{microsoft2025phi4mini}, under the resource constraints typically encountered in in-vehicle systems. In particular, the parameter sizes of the selected models do not exceed 7B, ensuring that they are feasible for real-world deployment. Our experimental results indicate that the mainstream models still perform suboptimally in all three sub-tasks. Across every task, the average scores remained below 60\%, underscoring the fact that current VLMs lack the specialized driving knowledge and situational understanding necessary to reliably handle complex traffic scenarios. Consequently, they fall short of meeting the stringent requirements demanded by high-reliability autonomous driving applications.

Furthermore, we introduce a new method named SafeDriveRAG, a plug-and-play retrieval-augmented generation (RAG) approach based on a multimodal graph structure to enhance the safety of autonomous driving system. This method transforms a large corpus of traffic safety documentation into a structured multimodal knowledge graph, utilizing textual, visual, and semantic entities. Together with our novel multi-scale subgraph retrieval algorithm, the approach enables efficient information extraction and enhanced inference, substantially improving the quality of generated responses and inference accuracy in safety-critical driving scenarios. Experimental results further confirm the effectiveness of our approach: across the three sub-tasks, the model’s performance improved by 4.73\%, 8.79\%, and 14.57\%, demonstrating substantial potential for real-world driving scenarios.

Our main contributions can be summarized as follows:

\par\textbf{(1)}  We propose the first large-scale autonomous driving safety QA benchmark, named \emph{SafeDrive228K}, which is a multitask and multi-scenario dataset for systematically evaluating the reasoning capabilities of widely-utilized VLMs in safety-critical driving contexts.

\par\textbf{(2)}  We design a new baseline \emph{SafeDriveRAG}, a multimodal graph-based plug-and-play RAG method that significantly enhances VLMs information utilization, reasoning and generation capabilities in traffic safety tasks, achieving substantial improvements over the original models across all three sub-tasks.

\par\textbf{(3)}  We conduct a systematic evaluation of multiple mainstream open-source VLMs, highlighting their performance and limitations in handling traffic accidents, corner cases, and traffic safety commonsense.

\section{Related Work}

\noindent\textbf{VLMs in Autonomous Driving.}~In recent years, advances in research on Vision-Language Models~\cite{openai2024gpt4,qwen2vl2024,8953263,yu2024rlaif, mao2024language,lv2024sgformer, gopalkrishnan2024multi} (VLMs) have led to significant progress in the joint comprehension of images and text, demonstrating outstanding performance on tasks such as image captioning, cross-modal retrieval, and visual question answering. Leveraging these versatile capabilities, researchers have begun integrating VLMs into autonomous driving systems to enhance safety and interpretability. For instance, LMDrive~\cite{shao2024lmdrive}  combines a visual encoder with a large language model, enabling natural language command execution within autonomous driving systems; 
and DriveVLM~\cite{tian2024drivevlm} introduces a Chain-of-Thought~\cite{wei2022chain} (CoT) inference mechanism that allows the VLM to generate a complete driving trajectory plan. However, most existing VLMs rely primarily on generic image-text data for pre-training and lack real-world driving experience or specialized domain knowledge, constraining their effectiveness in safety-critical scenarios. Consequently, in this work, we develop a newly multimodal question-answering benchmark focused on traffic safety encompassing tasks like traffic accidents, corner cases, and common safety knowledge, to promote the practical applicability of VLMs in real-world road environments.

\noindent\textbf{Multimodal Driving Datasets.}~As shown in Table~\ref{tab:dataset_comparison}, alongside the development of autonomous driving technologies, researchers have constructed various datasets endowed with vision-language capabilities~\cite{deruyttere2019talk2car,park2024vlaad}, targeting diverse perception and decision-making tasks. For instance, NuScenes-QA~\cite{qian2024nuscenes} and CODA-LM~\cite{chen2024automated} primarily address scene description, environmental perception, and driving advice generation; DriveLM~\cite{sima2025drivelm} employs a graph-structured question-answering paradigm to integrate perception, prediction, planning, action generation, and motion control; and IDKB~\cite{lu2024lvlms}  draws on extensive driving manuals and simulation data to establish a large-scale knowledge base for the traffic domain. Despite the substantial progress made by these multimodal datasets for perception and prediction tasks, they often neglect the systematic assessment of traffic safety considerations. Existing work~\cite{Xu_2020_CVPR,malla2023drama,9052709, tian2024drivevlm,xu2024drivegpt4,lv2023disentangled,9351755} tends to focus on granular technical evaluations, lacking a thorough examination of VLMs’ performance in acquiring safety-related knowledge, responding to risks, and performing complex reasoning. To fill this gap, in this work, we present the first large-scale multimodal question-answering benchmark centered on traffic safety comprehension and reasoning skills. In contrast to traditional datasets that focus on perception and planning, our benchmark emphasizes VLMs' ability to understand and reason about a wide range of hazardous and complex driving situations.

\noindent\textbf{Retrieval-Augmented Generation~(RAG).}~RAG can enhance large model outputs by retrieving relevant information from external knowledge sources ~\cite{ding2024retrieveneeds,gao2024retrievalaugmented}. 
Existing RAG methods generally fall into two categories: text-block-based approaches~\cite{mao2020generation,qian2024memorag} that segment text into minimal semantic units, parenthesis or sentences, to facilitate rapid matching, and graph-based approaches~\cite{edge2025localglobalgraphrag,guo2024lightrag,8954105,fan2025minirag} which extract entities and utilize structured modeling to build knowledge graphs that improve the semantic relevance of retrieval. As large language models expand to multimodal inputs, multimodal RAG~\cite{hu2023reveal,chen2022murag} has emerged, supporting joint retrieval and interpretation. Against this backdrop, we propose a graph-based multimodal RAG framework. Compared to previous graph-based methods, our approach employs a multi-scale subgraph retrieval algorithm to mitigate interference from redundant information and reduce computational overhead, thus improving entity and text-block matching efficiency. We also extend the types of retrievable information by introducing image nodes into the knowledge graph, further boosting the model’s capacity for knowledge integration and reasoning in complex traffic scenarios.

\section{SafeDrive228K Benchmark}

We propose the first large-scale benchmark \emph{SafeDrive228K}, which encompasses three domains, i.e., traffic accidents, corner cases, and traffic safety commonsense. The benchmark includes totally 228K multimodal question-answer pairs that aims to comprehensively evaluate models’ cognitive and reasoning capacities in diverse traffic safety scenarios. Below, we outline the benchmark’s construction process in detail.

\subsection{Data Source}

\noindent\textbf{Traffic Accidents.}~We incorporate a traffic accident subset into the SafeDrive228K benchmark to address the stringent safety requirements in accident scenarios, which offer a rigorous test of the perception and understanding in hazardous contexts. For this purpose, we adopt CAP-DATA~\cite{fang2022cognitive} as the primary data source, which contains 11K real-world accident videos with detailed annotations.

\noindent\textbf{Corner Cases.}~In real-world driving environments, corner cases generally refer to rare but high-risk scenarios, such as sudden animal crossings or visibility issues caused by extreme weather. Compared to traffic accidents, corner cases place greater emphasis on the model’s reasoning and response to `unknown hazards'. To build this subset, we adopt CODA-LM~\cite{chen2024automated}, which includes 9,768 images capturing from real-world edge-case road scenarios.

\noindent\textbf{Traffic Safety Commonsense.}~Finally, the traffic safety commonsense subset focuses on fundamental knowledge and rules of everyday driving, such as recognizing road signs and adhering to traffic legal regulations. In contrast to the two aforementioned subsets, this portion aims to assess models' `basic safety knowledge' and `daily compliance' competencies. Specifically, we collected approximately 1,100 relevant documents, amounting to over 2,600 pages from Chinese internet sources. We also incorporated driving manuals and test data from the IDKB dataset~\cite{lu2024lvlms} to maximize coverage of various traffic safety knowledge points. Since some content in IDKB is available in multiple languages, and most VLMs currently offer limited multilingual support, we converted the relevant data into English and merged it with the collected documents.

\begin{figure*}[t]
    \centering
    \includegraphics[width=0.9\textwidth]{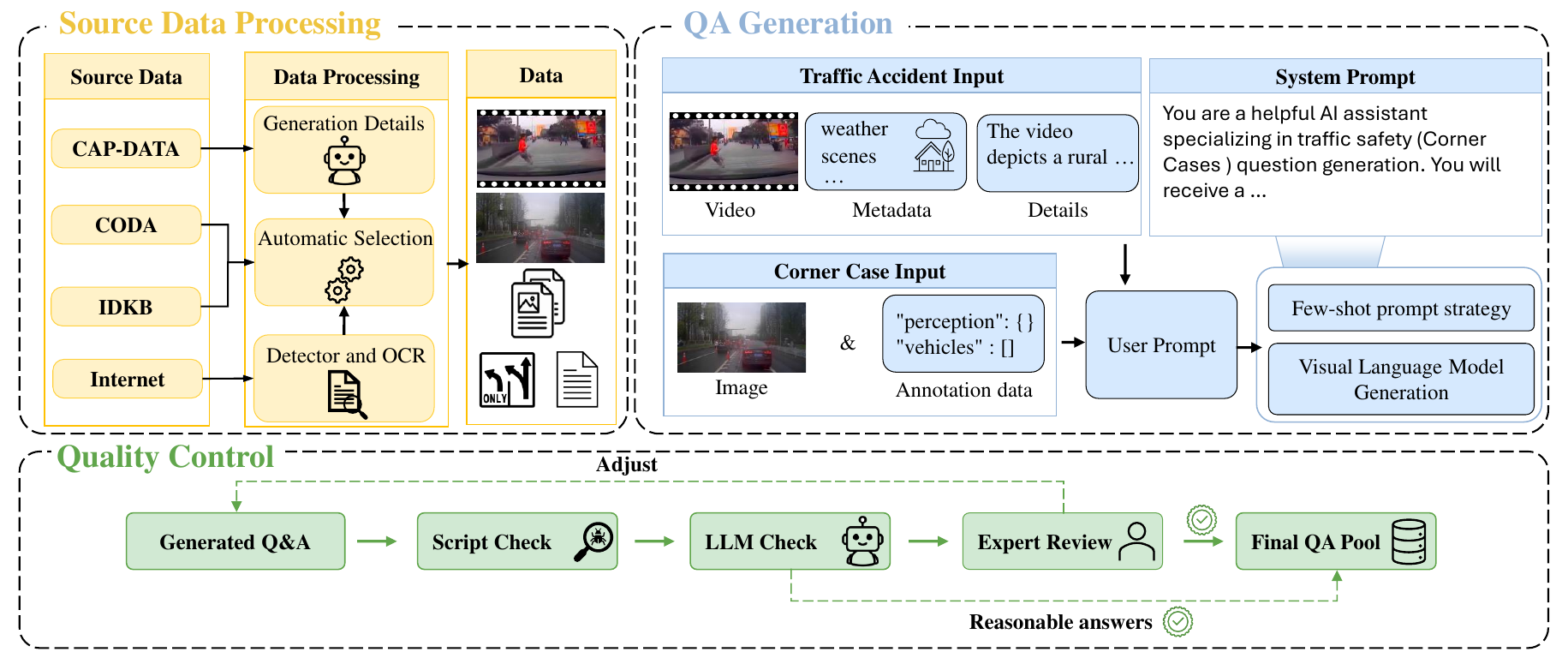}
    \vspace{-2mm}
    \caption{
    Our proposed SafeDrive228K benchmark is constructed through a semi-automated workflow that consolidates and process multiple source datasets, and then employs various tools with VLMs, with expert quality checks ensuring the reliability and accuracy of the final Q$\&$A. Then, the QA generation takes traffic accident and corner cases as input to synthesis prompts.
    }
    \label{fig:pipline}
    \vspace{-2mm}
\end{figure*}


\begin{figure}[tbp]
    \centering
    \begin{subfigure}[b]{0.23\textwidth}
        \centering
        \includegraphics[width=\linewidth]{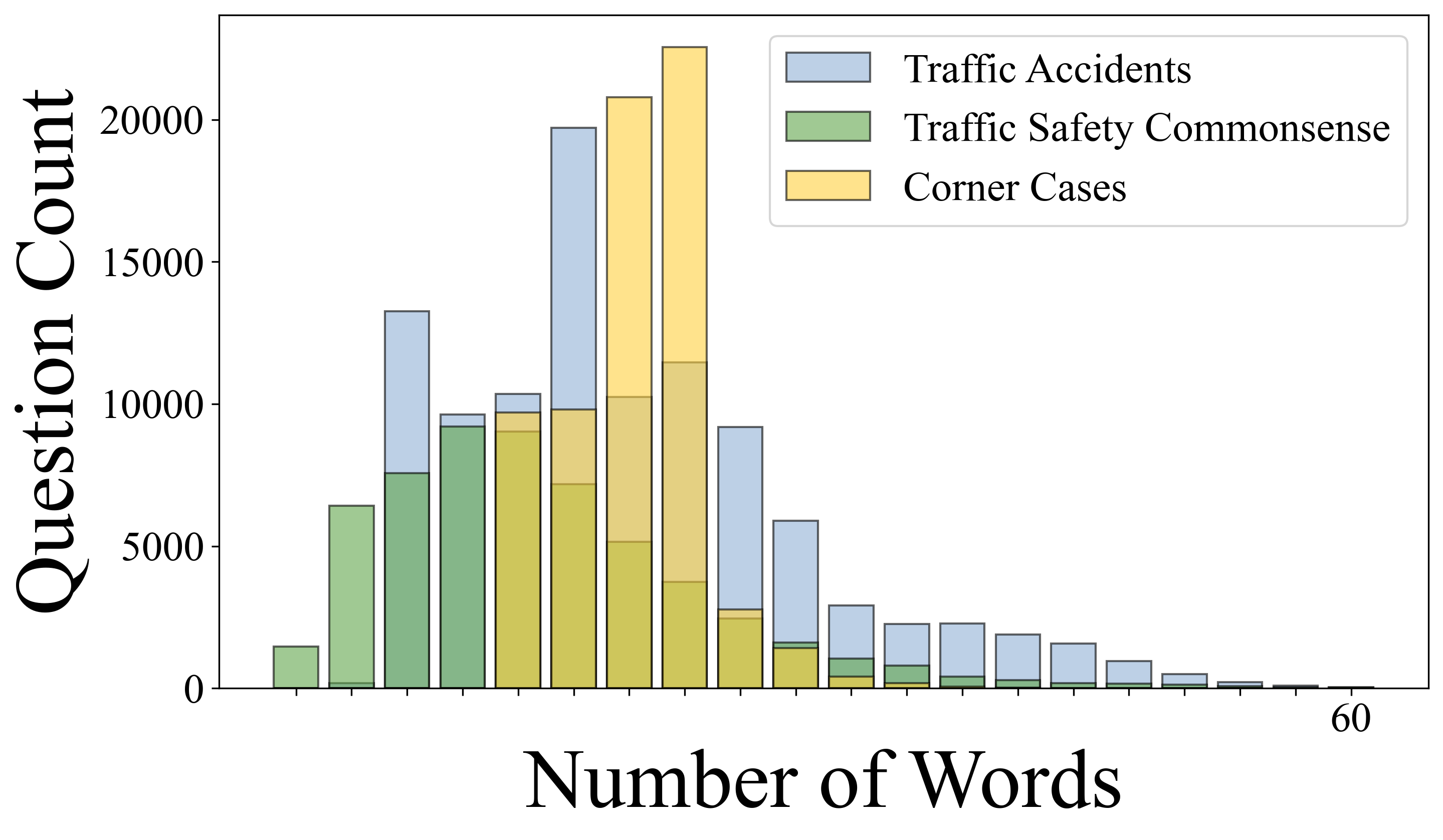}
        \caption{Distribution of question word across different sub-tasks.}
    \end{subfigure}
    \hfill
    \begin{subfigure}[b]{0.23\textwidth}
        \centering
        \includegraphics[width=\linewidth]{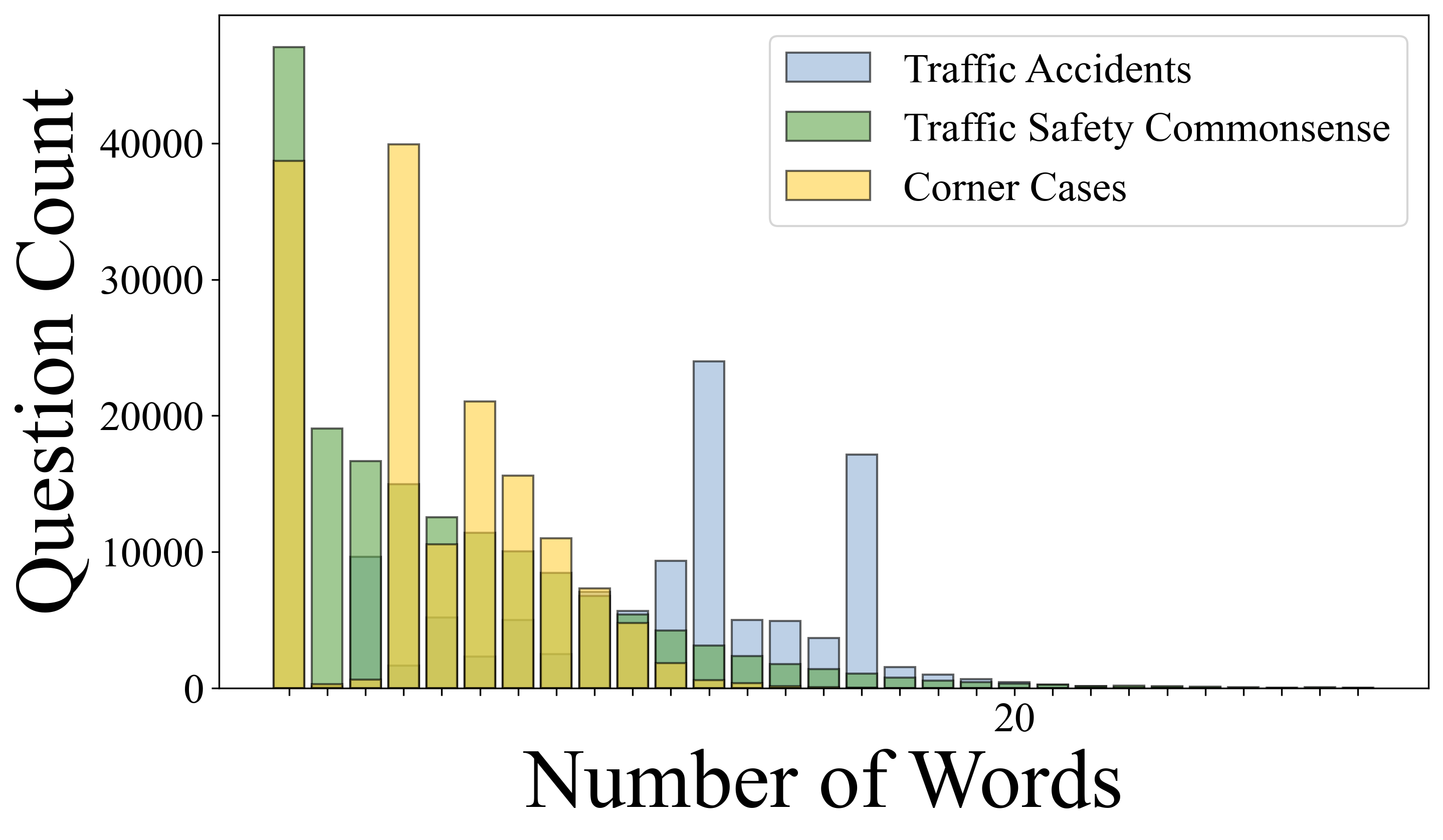}
        \caption{Distribution of MCQ word across different sub-tasks.}
    \end{subfigure}
    
    \vspace{0.1cm}
    \begin{subfigure}[b]{0.23\textwidth}
        \centering
        \includegraphics[width=\linewidth]{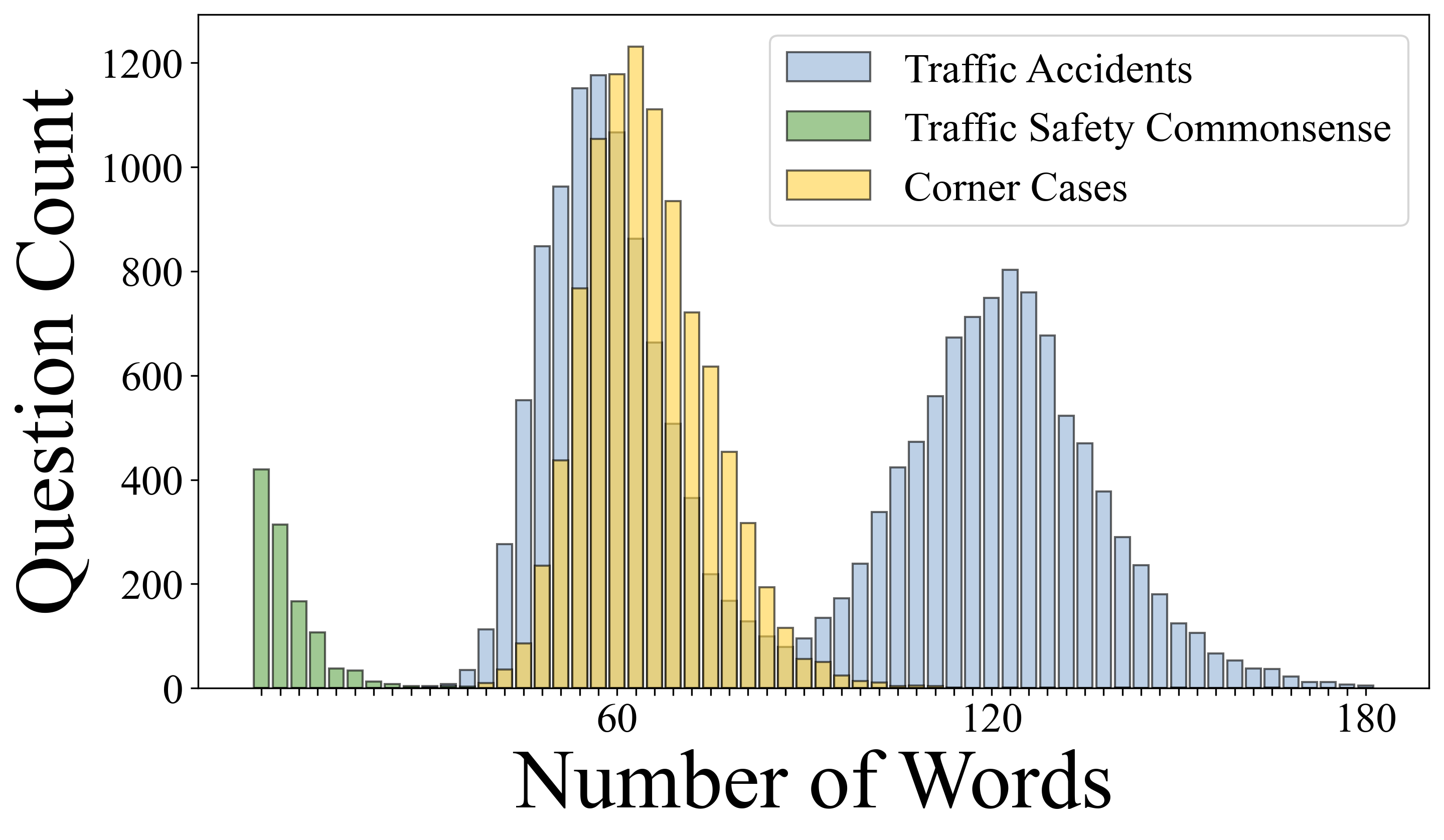}
        \caption{Distribution of  QA answers word across different sub-tasks.}
    \end{subfigure}
    \hfill
    \begin{subfigure}[b]{0.23\textwidth}
        \centering
        \includegraphics[width=\linewidth]{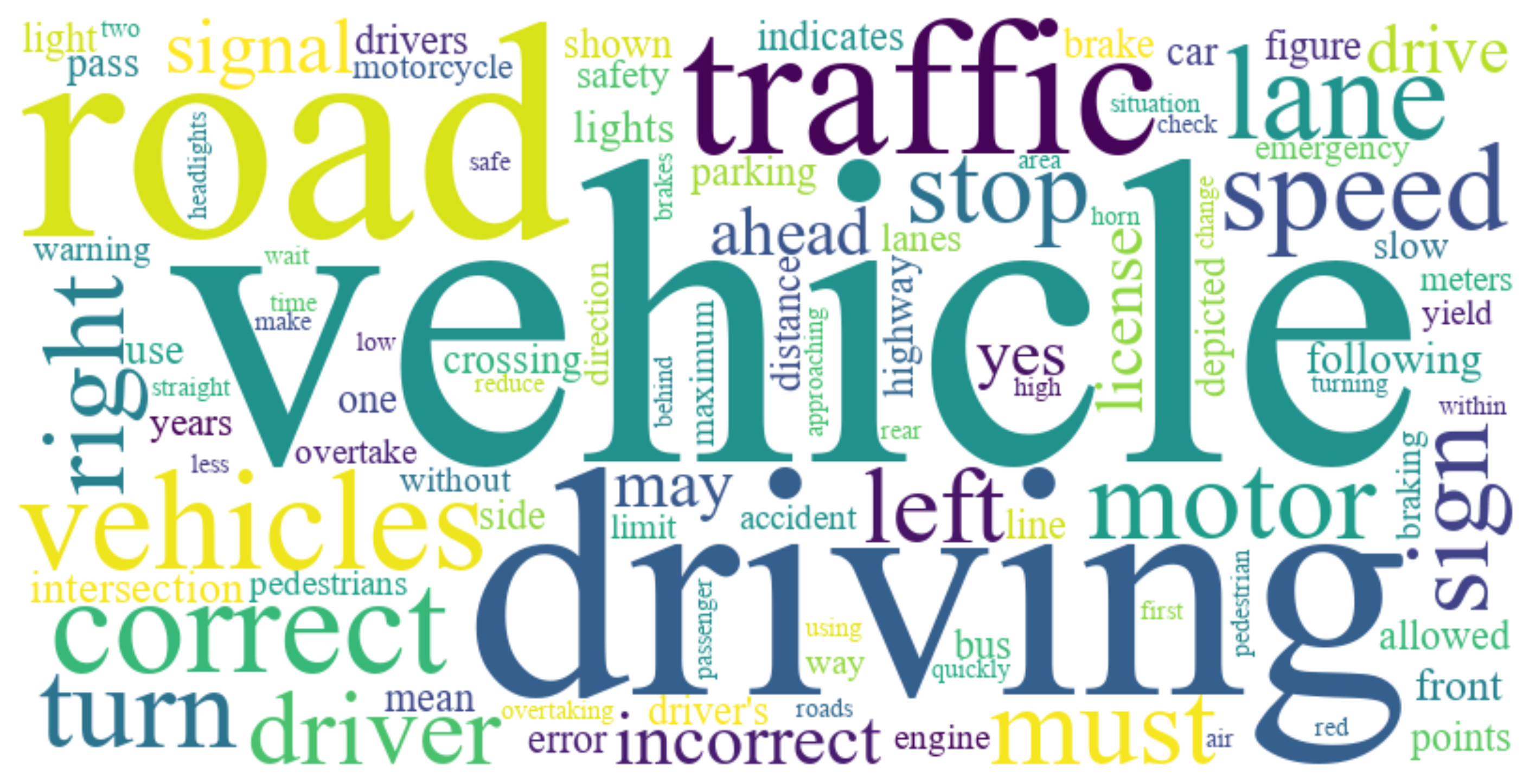}
        \caption{Word cloud distribution of all questions.}
    \end{subfigure}
    \vspace{-3mm}
    \caption{Statistical distributions of our proposed benchmark, \emph{i.e.}, distributions of word counts in questions and answers, and a top-k word cloud visualization of questions. MCQ means Multiple Choice Question.}
    \label{fig:distributions}
    \vspace{-3mm}
\end{figure}

\begin{figure*}[!t]
    \centering
    \includegraphics[width=0.86\textwidth]{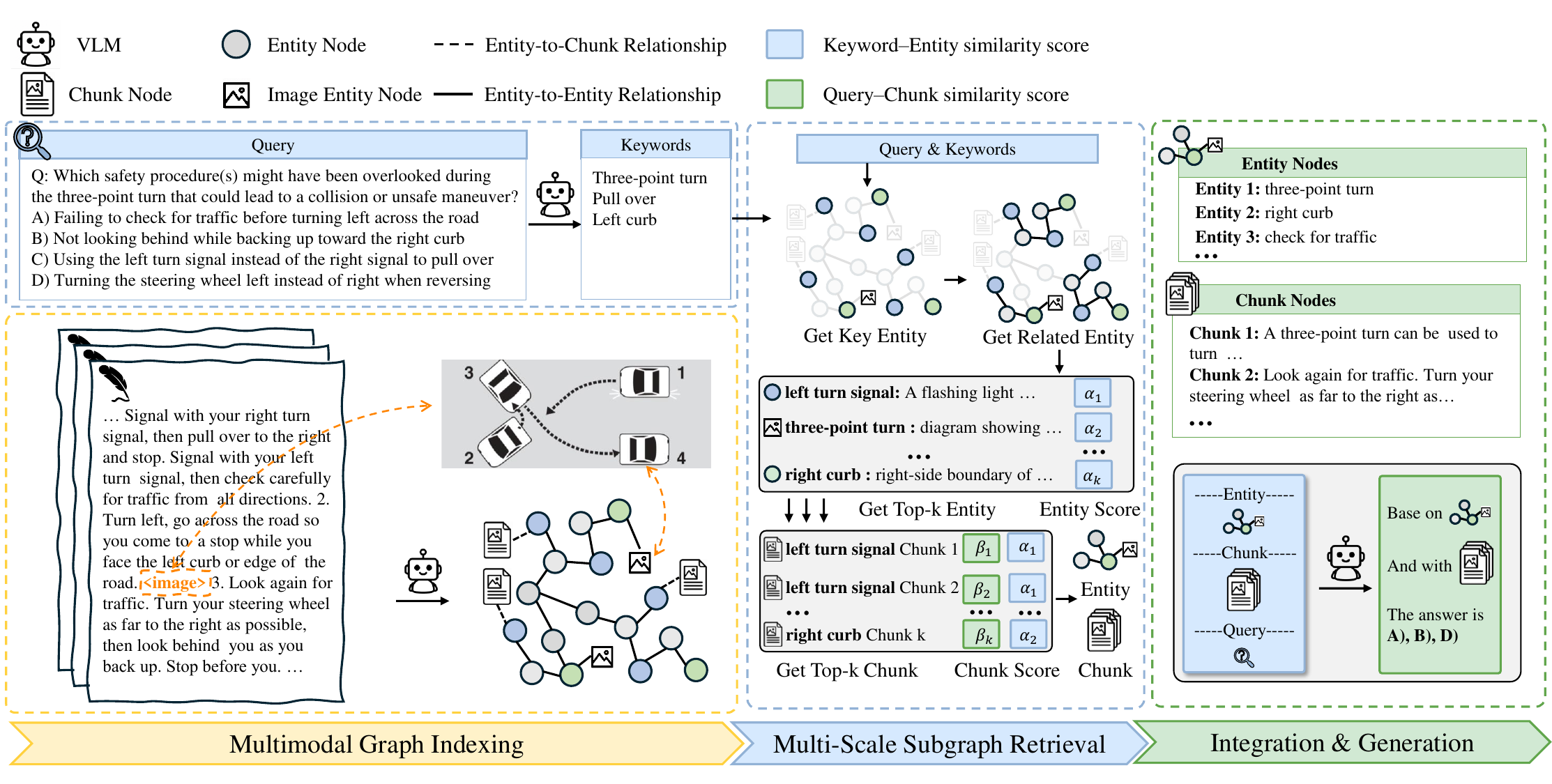}
    \vspace{-3mm}
    \caption{
    SafeDriveRAG leverages a multimodal graph indexing method in conjunction with a multi-scale subgraph retrieval approach, effectively overcoming the challenge of slow subgraph retrieval and enhancing both operational efficiency and overall performance.
    }
    \label{fig:architecture}
   \vspace{-3mm}
\end{figure*}

\subsection{Dataset Construction}

As depicted in Fig. \ref{fig:pipline}, we design a semi-automated data generation process, which comprises three main stages: source data processing, Q$\&$A pair generation, and data validation.

\noindent\textbf{Source Data Processing.}~For the traffic accident subset, we meticulously selected 9K videos from the original collection to form the basis of our benchmark, removing clips with poor image quality. To ensure that critical information in each video was fully captured during Q$\&$A generation, we leveraged a large language model~(LLM) to produce detailed video descriptions for each clip, building on CAP-DATA’s~\cite{fang2022cognitive} existing annotations. For corner cases subset, we selected 9K representative images from CODA-LM~\cite{chen2024automated} to focus on rare but high-risk road scenarios. With respect to traffic safety commonsense subset, we employed layout detection and OCR techniques~\cite{li2022ppocrv3} to extract text and data blocks from the collected documents, categorizing them into `Traffic Safety Commonsense Documents' and `Traffic Safety Commonsense Driving Test Documents'. We then reorganized the extracted information from the driving test documents into a standardized question-answer format. Lastly, we integrated the driving manuals and test data from IDKB~\cite{lu2024lvlms} into these two document sets. In this framework, the Traffic Safety Commonsense Documents serve as references for subsequent RAG tasks, whereas the Driving Test Documents are utilized for model evaluation.

\noindent\textbf{Q$\&$A Pair Generation.}~To ensure diversity in the generated questions, we adopted a few-shot prompt strategy. For each question type, we designed a corresponding prompt and provided at least one example for each distinct question category. After generating the Q$\&$A pairs for each category, we performed a preliminary inspection to verify the correctness of the output format. Any questions or answers that did not adhere to the required format were regenerated. 

\noindent\textbf{Data Quality Control.}~To guarantee high-quality outputs, we conducted a three-step validation procedure on the generated Q$\&$A pairs. First, we employed scripts to filter out obviously incorrect entries, such as empty answer options or overly short responses. Next, we used GPT-4o-mini~\cite{openai2024gpt4} to evaluate the remaining answers for logical consistency. Finally, for questions flagged as containing unreasonable answers, we enlisted 10 experts with driving experience to perform an in-depth review, providing professional answers for contentious items. These expert-reviewed answers served as our gold-standard data to ensure reliability and accuracy.

Following the above process, we obtained totally 228K Q$\&$A pairs of high quality that span multiple driving scenarios, providing a solid foundation for subsequent evaluations of VLM performance in traffic safety–oriented autonomous driving tasks.

\subsection{Dataset Statistics}

As shown in Table~\ref{tab:dataset_comparison}, our benchmark offers three primary innovations compared with existing autonomous driving datasets.

\noindent\textbf{Traffic Accidents.}~A unique aspect of this benchmark is our inclusion of real-world traffic accident videos as one of its core data sources for question generation. For accident scenarios, we designed 7 targeted question types, covering \emph{accident type detection}, \emph{accident prevention}, \emph{emergency handling}, \emph{legal regulation recognition}, \emph{right-of-way determination}, \emph{complex road condition decision-making}, and \emph{general questions}. General question here refers to allowing the model to independently generate questions, without imposing any specific question bias or constraint. This approach ensures broader coverage, thereby enhancing the assessment of the model’s overall understanding. This design aims to comprehensively evaluate a model’s capabilities in managing high-risk scenarios, spanning the full timeline from pre-accident risk prediction to post-accident response. Based on 9,331 traffic accident videos, we automatically generated a total of 102K Q$\&$As. 

\noindent\textbf{Corner Cases.}~Our benchmark also encompasses numerous object-level corner cases to evaluate a model’s safety performance when handling rare and complex scenarios. In the corner case, we devised 5 categories of questions: \emph{object recognition}, \emph{object localization}, \emph{danger prevention}, \emph{emergency handling}, and \emph{general questions}. These questions focus on a model’s perception, reasoning, and response under 'unknown hazard' conditions. Drawing on 9,768 corner case scenarios, we ultimately generated 69K questions, providing a rich resource to test VLM performance in atypical driving contexts. 

\noindent\textbf{Traffic Safety Commonsense.}~To facilitate a comprehensive assessment of a model’s understanding of everyday driving safety knowledge, we collected an extensive set of traffic safety commonsense data from both domestic and international sources. After applying layout detection and OCR for structured data extraction and standardized processing, we compiled and generated 57K Q$\&$A items. These cover a wide range of topics including driving regulations, traffic sign recognition, and behavioral guidelines.

Ultimately, our benchmark in total comprises 9,331 traffic accident videos and 35K images, for a total of 228K multimodal Q$\&$A pairs. As illustrated in Fig. \ref{fig:distributions}, we show several statistical distributions, such as question complexity, answer length, and a keyword cloud, to provide a more intuitive overview of the dataset’s content characteristics and coverage.

\section{SafeDriveRAG Method}

Most existing Vision-Language Models (VLMs) are pre-trained on general-purpose datasets and lack specialized domain knowledge in traffic safety~\cite{tian2024drivevlm}. In real-world autonomous driving scenarios,particularly those that are complex or pose high risks—this deficiency can lead to significant safety concerns, as the model may struggle to make accurate judgments. To address this issue, we propose a new Graph-RAG framework, namely~\emph{SafeDriveRAG}, which not only extends the types of information retrievable within the graph structure but also introduces an efficient multi-scale subgraph retrieval algorithm. By substantially reducing retrieval latency, it enhances inference efficiency for autonomous driving, where rapid response times are critical.

The overall architecture of our proposed RAG framework is depicted in Fig. ~\ref{fig:architecture} and comprises two core modules: \emph{Multimodal Indexing Module} that constructs a semantically aware representation of multimodal knowledge, and \emph{Multi-Scale Subgraph Retrieval Module} that enables fast and accurate information retrieval.

\subsection{Problem Setting}
We formally define the SafeDriveRAG task as a multimodal question-answering problem. Given an input scene $I$, which may be a video $V$ or an image $S$,  together with an associated question $Q$, the objective is to predict the most suitable answer or generate relevant text in response. For multiple-choice questions, each question $Q$ is accompanied by a finite set of candidate answers $A=\{a_1,a_2,\cdots,a_n\}$. The goal is to select one or more options from $A$ that best address $Q$. The formulation can be expressed as follows:
\begin{equation}a^*=\arg\max_{a\in{A}}vlm(a| I,Q),\end{equation}
where $vlm$ refers to the VLM model, and $a^*$ denotes the chosen optimal subset of answers. For open-ended question-answering, the objective is to generate a natural language answer $A$ with the highest semantic relevance, based on the input modalities (video or image) and the question $Q$. This can be written as:
\begin{equation}A=vlm(Q, I).\end{equation}

On the other hand, we define the RAG framework, denoted by $M$. In RAG, the indexing module $R$ is responsible for extracting a knowledge graph $G$  from the raw documents and constructing an index $\phi$. The retrieval module $\psi(\cdot)$ subsequently leverages the query and indexed data to locate the relevant documents. Formally:
\begin{equation}{M}=\begin{pmatrix}{G},{R}=(\varphi,\psi)\end{pmatrix}.\end{equation}

\subsection{Multimodal Indexing Module}~In real-world traffic scenarios, most safety documents contain rich textual data alongside images, charts, and other visual elements. Existing RAG systems typically support only text-level knowledge retrieval, which limits their ability to effectively utilize visual content. To enhance safe decision-making in realistic driving environments, we designed a multimodal indexing module tailored to traffic safety contexts.

Within this module, we build a heterogeneous graph structure ${G=\{V,E\}}$ containing three types of nodes and two types of edges, providing a unified representation of both text and images, as the following:

\noindent\textbf{Entity Node} ${V}_{e}$: To capture structured knowledge, represent key semantic units in the documents.

\noindent\textbf{Image Entity Node} ${V}_{i}$: Correspond to important image content referenced in the documents. We insert a placeholder symbol \mbox{<image>} in the text to indicate the location of the image.

\noindent\textbf{Text Chunk Node} ${V}_{c}$: Represent original contextual paragraphs, preserving the complete semantic information needed for contextual modeling.

\noindent\textbf{Entity–Entity Edge} ${E}_{ee}$: Capture semantic relationships or spatial/temporal associations between entities.

\noindent\textbf{Entity–Chunk Edge} ${E}_{ec}$: Link entities to their corresponding contextual fragments, ensuring semantic coherence and traceability.

Ultimately, the output of the indexing module is a heterogeneous multimodal knowledge graph ${G}$, which serves as a unified and enriched representation of traffic safety knowledge, as follows:
\begin{equation}
    {G}=(\{{V}_e,{V}_i,{V}_c\},\{{E}_{ee},{E}_{ec}\}).
\end{equation}

\begin{table*}[t]
\small
  \caption{Quantitative results of VLMs across all tasks (i.e., traffic accidents, corner cases and traffic safety commonsense). Green indicates the methods incorporating our proposed RAG method. Bold highlights the better results.}
  \vspace{-3mm}
  \label{tab:results}
  \centering
  \begin{tabular}{@{}l|l |c c c|c c c c |c@{}}
    \toprule
    & & \multicolumn{3}{c}{\textbf{Multi-choice Question}} 
      & \multicolumn{4}{c}{\textbf{Question \& Answer}} 
      & \multirow{2}{*}{\textbf{SafeDrive Score}} \\
    \cmidrule(lr){3-5} \cmidrule(lr){6-9}
    \textbf{Models} & \textbf{Category} 
                    & \textbf{Single} & \textbf{Multiple} & \textbf{Overall}
                    & \textbf{R-1} & \textbf{R-L} & \textbf{SemScore} & \textbf{Overall} 
                    & \\
    \midrule
    \multirow[c]{4}{*}{\raisebox{-0.5ex}{\shortstack[l]{\textbf{Qwen-2.5-vl~\cite{bai2025qwen25vl}}}} }
& 3B w/o rag  
         & 54.91\% & 46.20\% & 54.23\%  
         & 15.31\% & 14.01\% & 51.33\% & 26.88\% & 44.89\% \\
    & \cellcolor{green}3B w/ rag   
        & \cellcolor{green}\textbf{65.43}\% 
        & \cellcolor{green}\textbf{48.93}\% 
        & \cellcolor{green}\textbf{62.50}\%  
        & \cellcolor{green}\textbf{19.25}\% 
        & \cellcolor{green}\textbf{16.99}\% 
        & \cellcolor{green}\textbf{55.33}\% 
        & \cellcolor{green}\textbf{30.52}\% 
        & \cellcolor{green}\textbf{57.38}\% \\

        \cmidrule(lr){2-10}
        
    & 7B w/o rag  
         & 61.42\% & \textbf{58.70}\% & 61.90\%  
         & 16.57\% & 14.91\% & 56.79\% & 29.42\% & 53.14\% \\
    & \cellcolor{green}7B w/ rag   
        & \cellcolor{green}\textbf{67.81}\% 
        & \cellcolor{green}58.17\% 
        & \cellcolor{green}\textbf{66.07}\%  
        & \cellcolor{green}\textbf{20.05}\% 
        & \cellcolor{green}\textbf{18.32}\% 
        & \cellcolor{green}\textbf{58.04}\% 
        & \cellcolor{green}\textbf{32.14}\% 
        & \cellcolor{green}\textbf{60.18}\% \\
    \midrule
    \multirow[=]{4}{*}{\raisebox{-0.5ex}{\shortstack[l]{\textbf{LLAVAA-OneVision~\cite{li2024llavaonevision}}}}}
    & 0.5B w/o rag 
         & 22.90\% & \textbf{13.18}\% & 20.27\%  
         & 8.14\% & 7.25\% & 44.85\% & 20.08\% & 20.57\% \\
    & \cellcolor{green}0.5B w/ rag   
        & \cellcolor{green}\textbf{36.51}\% 
        & \cellcolor{green}10.89\% 
        & \cellcolor{green}\textbf{29.71}\%  
        & \cellcolor{green}\textbf{15.88}\% 
        & \cellcolor{green}\textbf{14.30}\% 
        & \cellcolor{green}\textbf{46.79}\% 
        & \cellcolor{green}\textbf{25.65}\% 
        & \cellcolor{green}\textbf{29.52}\% \\
        \cmidrule(lr){2-10}
    & 7B w/o rag   
         & 57.51\% & 38.25\% & 54.15\%  
         & 14.08\% & 13.04\% & 51.90\% & 26.34\% & 48.32\% \\
    & \cellcolor{green}7B w/ rag   
        & \cellcolor{green}\textbf{65.50}\% 
        & \cellcolor{green}\textbf{42.49}\% 
        & \cellcolor{green}\textbf{59.53}\%  
        & \cellcolor{green}\textbf{20.78}\% 
        & \cellcolor{green}\textbf{19.34}\% 
        & \cellcolor{green}\textbf{56.67}\% 
        & \cellcolor{green}\textbf{32.26}\% 
        & \cellcolor{green}\textbf{57.24}\% \\
    \midrule
    \multirow[=]{2}{*}{\shortstack[l]{\textbf{Phi-4-multimodal-instruct~\cite{microsoft2025phi4mini}}}}
    & 4.5B w/o rag
         & 37.88\% & \textbf{48.23}\% & 41.02\% 
         & 15.45\% & 14.18\% & 53.90\% & 27.84\% & 43.16\% \\
    & \cellcolor{green}4.5B w/ rag   
        & \cellcolor{green}\textbf{51.42}\% 
        & \cellcolor{green}45.19\% 
        & \cellcolor{green}\textbf{51.41}\%  
        & \cellcolor{green}\textbf{18.20}\% 
        & \cellcolor{green}\textbf{16.78}\% 
        & \cellcolor{green}\textbf{56.63}\% 
        & \cellcolor{green}\textbf{30.53}\% 
        & \cellcolor{green}\textbf{51.55}\% \\

    \bottomrule
  \end{tabular}
  \label{tab:ALL task}
  \vspace{-2mm}
\end{table*}

\begin{figure*}[t]
    \centering
    \includegraphics[width=0.87\textwidth]{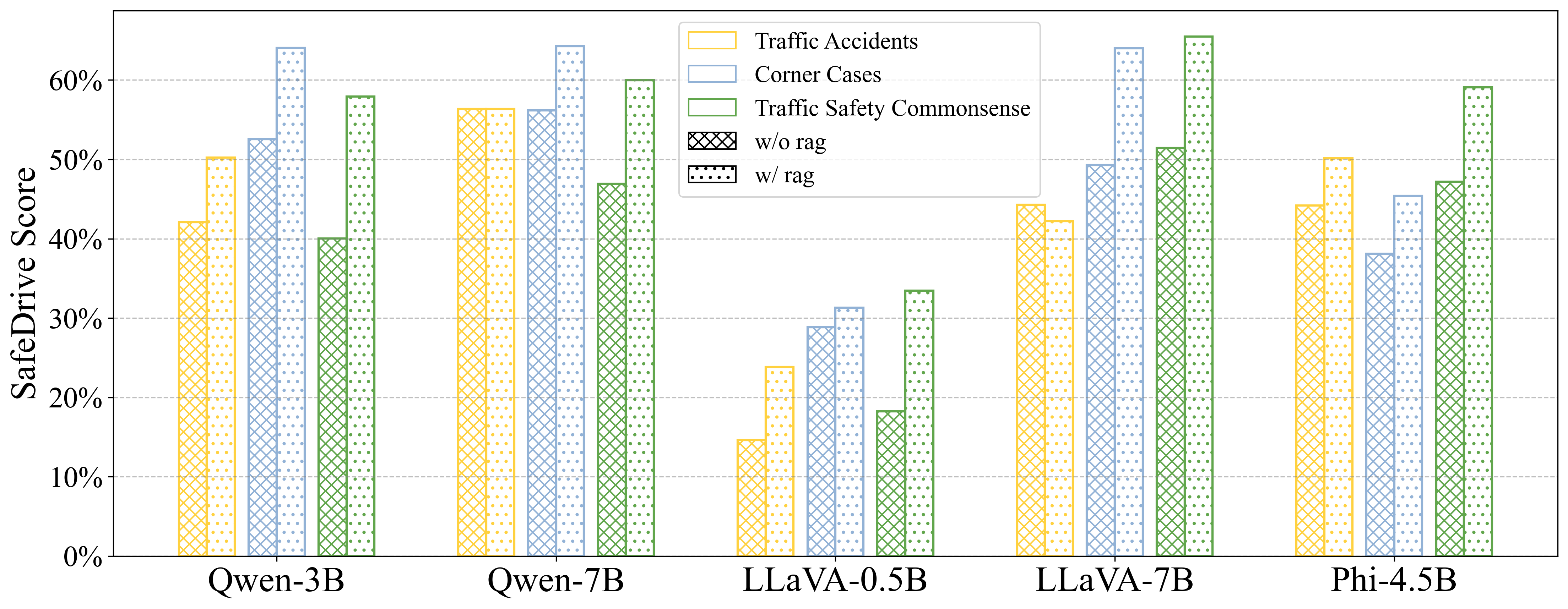}
    \vspace{-3mm}
    \caption{Results of the SafeDrive228K benchmark w.r.t SafeDrive Score across the sub-tasks of Traffic Accidents, Corner Cases, and Traffic Safety Commonsense.}
    \label{fig:histogram}
    \vspace{-3mm}
\end{figure*}

\subsection{Multi-Scale Subgraph Retrieval Module}

During the retrieval stage, our goal is to rapidly identify entities or chunks most pertinent to the input query, thereby supplying support for subsequent answer generation. The overall retrieval process proceeds as follows:

1.	Key Information Extraction: A Vision Language Model (VLM) is used to extract a set of critical keywords $K_Q = \{k_1,k_2,\ldots,k_m\}$ from  the input query $Q$, capturing its semantic focus.

2.	Multi-Scale Subgraph Retrieval: Using the extracted keywords $K_Q$ to obtain a semantic subgraph $G_Q\subseteq G$. 

3.	Answer Generation: The retrieved entities ${V}_e$, images ${V}_i$, and chunks ${V}_c$ are then passed to the Vision Language Model for the final answer production.

Most existing subgraph retrieval modules feature complex designs, often relying on extensive graph traversal or intricate path-matching algorithms. Although these approaches can perform well in offline knowledge reasoning, they struggle to fulfill the demanding real-time requirements of autonomous driving. To address this gap, we propose a multi-scale subgraph retrieval module that evaluates both entity-level and text-level semantics across multiple scales, thereby maintaining high retrieval quality while significantly improving efficiency and real-time inference. The module’s key design elements for achieving efficient retrieval are as follows:

1. Keyword-Driven Entity Initialization: Based on the keywords extracted from the query, we locate corresponding entity nodes in the graph by similarity scores, the formula is as follows: 
\begin{equation}{V}_{\mathrm{anchor}}=\{v_j\mid s(k_i,v_j)\geq\delta_1\},\end{equation}
where $\delta_1$ is a similarity threshold, $k_i \in K_Q$, $s(k_i,v_j)$  represents the semantic similarity between the keyword $k_i$ and entity $v_j$.

2. Multi-Hop Entity Expansion: For each anchor entity $v\in{V}_{\mathrm{anchor}}$, we perform $h$-hop expansion to obtain candidate entities ${V}_{cand}$:
\begin{equation}{V}_{cand} = \bigcup_{v \in V_{anchor}} Expand(v,h).\end{equation}
Starting from the anchor points, we perform multi-hop semantic expansion within the graph, bounded by a preset hop limit to discover potentially relevant entity nodes.

3. Entity Semantic Relevance Matching: For each expanded entity node, we calculate its semantic similarity with the query’s keywords and select the top-k entities.

4. Chunk Semantic Matching: Drawing on the textual blocks associated with the selected entity nodes, we compute the similarity $S(c)$ between the query $q$ and each chunk $c$:
\begin{equation}S(c)=\alpha\cdot\left(\sum_{v\in V_{cand}}s(q,v)\cdot\lambda^{k_{v}}\right)+(1-\alpha)\cdot{s(q,c)},\end{equation}
where $\lambda$ is the path decay factor, $k_v$ is the number of hops, and $\alpha$ is the chunk semantic score weight. 

5. Multimodal Information Output: Finally, the chosen entities $V=\{v_1,...,v_k\}$, along with their associated images and chunks $C=\{c_1,...,c_k\}$, are combined into a high-relevance subgraph. This subgraph is then fed into the VLM to provide coherent and context-rich knowledge for the answer-generation phase.

Through this design, our Graph-RAG achieves an optimal balance between efficiency and accuracy, offering a stable, rapid, and scalable retrieval-augmented capability ideally suited for real-world applications such as autonomous driving.

\section{Experiments}
In this section, we present a systematic evaluation of five mainstream open-source VLMs on our newly constructed multimodal traffic safety question-answering benchmark. These experiments aim to assess each model’s understanding and reasoning capabilities under safety-critical driving scenarios.

\subsection{Experimental Settings}
\textbf{Model Select}: We selected five mainstream open-source VLMs, all with parameter sizes under 7B, to accommodate the stringent computational constraints of in-vehicle deployment. Compared to larger-scale models, lightweight VLMs exhibit higher adaptability in real-world scenarios and more accurately reflect the actual performance achievable in an on-board environment. Specifically, we include the following models: Qwen2.5-vl-7B~\cite{qwen2vl2024}, Qwen2.5-vl-3B~\cite{qwen2vl2024}, LLAVAA-OneVison-7B~\cite{li2024llavaonevision}, LLAVAA-OneVison-0.5B~\cite{li2024llavaonevision}, Phi-4-multimodal-instruct-4.5B~\cite{microsoft2025phi4mini}. Each model was evaluated in both its original setting and a RAG-enhanced variant. To ensure fairness, we used a uniform prompt across all models under the same input conditions. It should be noted that RAG-enhanced versions only incorporate external knowledge, such as entities and chunks, to assist models in understanding and generating appropriate responses.

\noindent\textbf{RAG Configuration}: To enhance the knowledge support of VLMs in traffic safety scenarios, we introduced a retrieval-augmented generation (RAG) mechanism and constructed an extensive external document repository. This repository primarily consists of traffic safety legislation, driving guides, and accident response manuals collected from the Internet, totaling over 1.32 million tokens. During text pre-processing, and following standard practices~\cite{guo2024lightrag}, we divided the documents into chunks with a maximum length of 1200 tokens, applying an overlapping sliding window of 100 tokens to preserve semantic integrity and contextual coherence. We then employed NanoVectorBase~\cite{nano_vectordb} to build an efficient vector-retrieval index, facilitating low-latency, semantically aware subgraph information retrieval. In the RAG module, the top-k for entity-node retrieval is set to 5, while the top-k for chunk retrieval is set to 3, balancing both entity- and context-level information for optimal performance.

\begin{figure*}[t]
    \centering
    \includegraphics[width=0.9\textwidth]{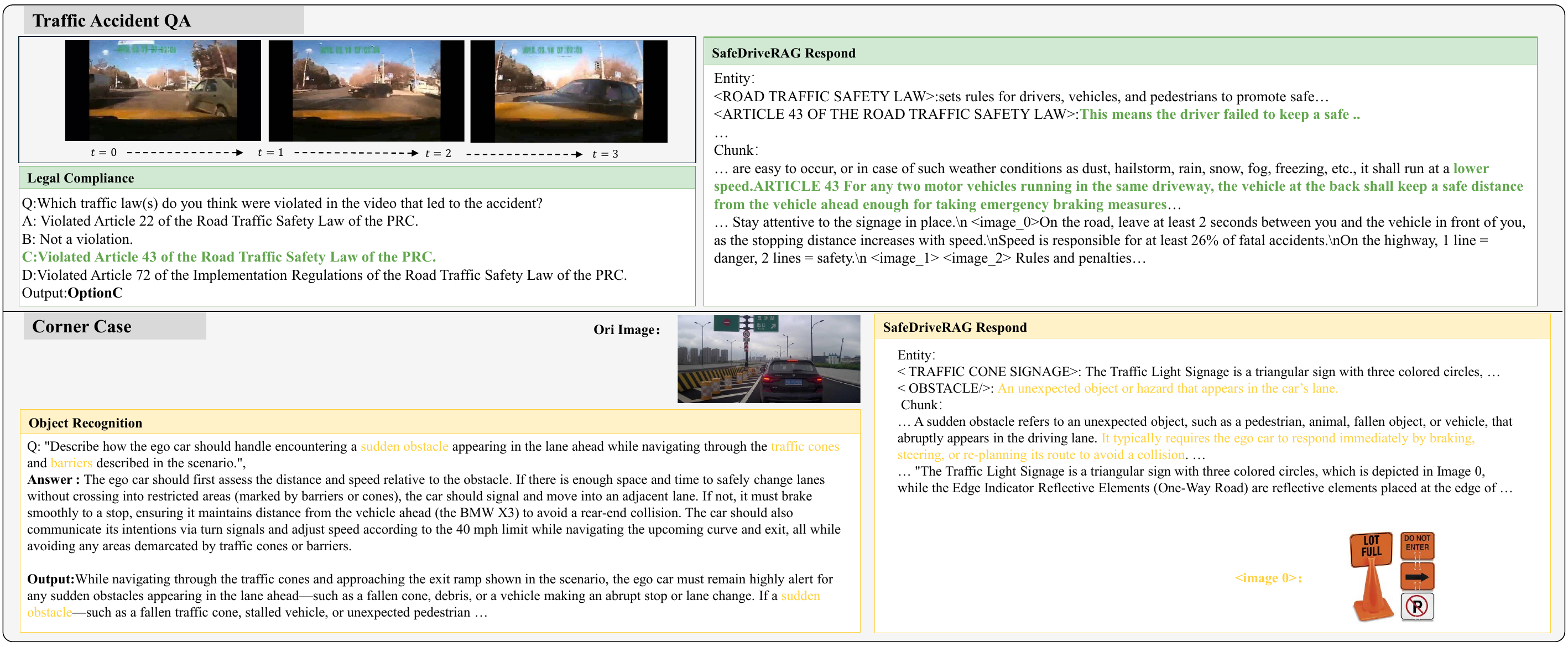}
    \vspace{-4mm}
    \caption{Visualized results of traffic accidents and corner cases, showing the retrieved entities, images, and chunks after subgraph retrieval.
    }
    \label{fig:visualization}
   \vspace{-3mm}
\end{figure*}

\subsection{Evaluation Metrics}
Throughout the evaluation, we divided the questions into two main types: \emph{multiple-choice} and \emph{open-ended}. Each type was assessed using different strategies: For multiple-choice tasks, we used regular expressions to extract the selected answer options from each VLM’s output and compared these options to the ground truth. To maintain rigorous standards, a model was deemed correct only if it provided all valid options for a given question. For open-ended tasks, we employed both \emph{ROUGE}~\cite{lin2004rouge} and \emph{SEMScore}~\cite{aynetdinov2024semscore} to measure the similarity between the model-generated answer and the reference answer. \emph{ROUGE} evaluates on N-gram overlap, while \emph{SEMScore} assesses sentence-level vector similarity to assess the semantic coherence of the generated answer, placing more emphasis on meaning rather than exact wording. We separately computed the scores for multiple-choice and open-ended questions, weighting each by the number of corresponding items. The results were then summed to yield an overall performance metric, referred to as the \emph{SafeDrive Score}. This composite score serves as a key indicator of how effectively a VLM performs in the safety-critical domain of autonomous driving. 

\subsection{Experimental Results}


\noindent\textbf{Overall Model Performance.} We presents the performance of various models in three sub-tasks within the SafeDrive228K, as summarized in Table \ref{tab:ALL task} and Fig. \ref{fig:histogram}. We can clearly observe that the overall performance of the original models on all tasks is generally poor. Across the three sub-tasks, most models fail to exceed 50\% on the SafeDrive Score, with only the Qwen-2.5-vl-7B model slightly surpassing 50. This finding underscores the limited modeling capacity and reasoning ability of current VLMs for traffic safety scenarios. A closer examination of each sub-task reveals that the traffic accident task yields the weakest results, with an average score of just 40.31\%. Performance on the corner case task averages 45.02\%, whereas the traffic safety commonsense task sees an average of 40.71\%. These findings highlight the substantial room for improvement in mainstream VLMs when dealing with traffic safety applications. Further enhancements involving structural optimization and knowledge augmentation appear necessary to overcome these challenges.

\noindent\textbf{Effectiveness of RAG Enhancement.} A comparison of pre- and post-RAG results in Fig. \ref{fig:histogram} shows that incorporating RAG significantly boosts the accuracy and completeness of model outputs. In particular, the average score increase in the traffic safety commonsense sub-task was 14.57, while corner cases improved by 8.79 points and traffic accident tasks by 4.73. Although there was a notable performance gain in the accident sub-task, the relatively small increase can be attributed to input truncation issues arising from large token counts (due to video data combined with entity and chunk information) in smaller models ($\leq$7B). Nonetheless, the example of Qwen2.5-vl-3B demonstrates that smaller models can approach or match 7B-level performance once equipped with RAG, indicating significant potential for lighter models under enhanced retrieval settings. 

\noindent\textbf{Discussion.} Overall, the Qwen series stood out, with Qwen2.5-vl-7B achieving an average score of 60.2, exhibiting greater consistency and stability across all tasks compared to other models. LLAVA-OneVision-7B showed the largest performance jump (average +14.39) after RAG integration in both corner cases and traffic safety commonsense tasks, suggesting a strong synergy with the retrieval mechanism. The Phi model, while smaller in size, still delivered robust results. In general, 7B models consistently outperformed their 3B counterparts, affirming the importance of model capacity in handling complex tasks.

\subsection{Ablation Study of RAG}

We conducted the ablation study on the traffic safety commonsense sub-task, and we adopted Qwen2.5-vl-7B with a 10\% test set to compare three widely-used RAG methods: Naïve RAG~\cite{mao2020generation}, MiniRAG~\cite{fan2025minirag}, and our proposed SafeDriveRAG. We report their measured subgraph retrieval times (Speed Time) were 46.22 s, 9519.98 s, and 884.10 s, respectively, while the SafeDrive Scores were 60.18\%, 61.26\%, and 62.07\%. Although Naïve RAG is the fastest in raw retrieval time, its lack of structured modeling leads to weaker overall performance; while MiniRAG achieves a modest accuracy boost but at the cost of significantly longer retrieval times. In contrast, our proposed SafeDriveRAG strikes a more favorable trade-off balance between efficiency and accuracy, demonstrating strong potential for real-world safety-critical deployments.

\section{Conclusion}

In this paper, we presented SafeDrive228K, the first comprehensive multimodal benchmark for evaluating VLMs in complex safety-critical driving scenarios. Furthermore, we designed SafeDriveRAG, a plug-and-play RAG method underpinned by a multimodal graph structure and an efficient multi-scale subgraph retrieval algorithm. Experiments showed that SafeDriveRAG consistently enhances performance across multiple safety-related driving tasks, demonstrating strong real-world deployment potential and promising contributions to related fields.

\begin{acks}
This work is partly supported by the Funds for the National Natural Science Foundation of China under Grant 62202063 and U24B20176, Beijing Natural Science Foundation (L243027), Beijing Major Science and Technology Project under Contract No. Z231100007423014.
\end{acks}

\bibliographystyle{ACM-Reference-Format}

\bibliography{acmmm}

\appendix

\end{document}